\documentclass[letterpaper]{article}

\usepackage[preprint]{aaai2027}
\usepackage[hyphens]{url}
\usepackage{graphicx}
\usepackage{natbib}
\usepackage{caption}
\usepackage{booktabs}
\usepackage{multirow}
\newcommand{\toolname}{\textsc{Terrarium}}

\title{\toolname: Growing Personal Assistants via Scalable Synthesis of App Worlds}
\author{
Mengzhou Wu\textsuperscript{1,2,\textdagger}, Yuzhe Guo\textsuperscript{1,2,\textdagger},
Jiawen Tao\textsuperscript{1}, Xiaokun Yuan\textsuperscript{1}\\
Yuan Cao\textsuperscript{1,2}, Gang Cao\textsuperscript{3},
Yuetang Deng\textsuperscript{3}, Zhi Gong\textsuperscript{3}\\
Dezhi Ran\textsuperscript{1,2,*}, Linyi Li\textsuperscript{4},
Wei Yang\textsuperscript{5}, Tao Xie\textsuperscript{1,2,5,*}
}
\affiliations{
\normalfont
$^{1}$School of Computer Science, Peking University, Beijing, China \\
$^{2}$Beijing Tongming Lake Information Technology Application Innovation Center, Beijing, China \\
$^{3}$WeChat, Tencent Inc., China \quad $^{4}$Simon Fraser University, Canada \\
$^{5}$Institute of Systems for Advanced Computing, Fudan University, China \\
$^{\dagger}$Equal contribution \quad $^{*}$Corresponding Authors
}

\begin{document}

\maketitle

\begin{abstract}
Advances in large language models (LLMs) have brought the vision of a general-purpose personal assistant closer to reality.
Such assistants could provide a unified natural-language interface to a broad ecosystem of applications.
However, real-world applications do not readily support training at scale because access is often restricted and their state is difficult to reset.
To address this challenge, we introduce \toolname, 
a framework that synthesizes executable App Worlds from real-world application information, generates training task packages with evaluator-side rubrics from these worlds, 
and iteratively validates and refines both through agent rollouts.
For App World synthesis, \toolname{} organizes public Web information and in-application observations into structured app specifications 
that guide the construction of self-contained runtimes with callable operations and mutable, resettable state. 
For task synthesis, \toolname{} constructs tasks grounded in App World state together with evaluator-side outcome rubrics that assess success without prescribing reference trajectories.
We construct a training suite of 275 App Worlds and 13{,}344 tasks for outcome-based reinforcement learning across three Qwen3 model scales (4B, 8B, and 14B parameters).
Across these scales, models trained with \toolname{} outperform the corresponding Qwen3 models by an average of 18.34 percentage points on the $\tau^2$-bench and 10.94 percentage points on the AppWorld bench.
These results suggest that app-grounded world and task synthesis offers a scalable path toward general-purpose assistants capable of operating across broad application ecosystems.
\end{abstract}

\section{Introduction}
\label{sec:introduction}

  Thousands of applications have become deeply integrated into daily life,
  creating an increasingly complex application ecosystem
  ~\citep{wen2024autodroid}.
A general-purpose personal assistant~\citep{muthusamy2023personalagents,liang2024taskmatrix} could provide a unified natural-language interface to this ecosystem.
Users express what they want to accomplish, and the assistant interacts with the relevant applications, plans the necessary operations, and carries out the task~\citep{qin2023toolllm,qin2025uitars}.
Advances in large language models (LLMs) have brought this vision closer to reality~\citep{openai2023gpt4,geminiteam2024gemini15,kimiteam2025kimik2}.
Training LLM agents across this ecosystem still depends on realistic, controllable, and scalable environments~\citep{zhou2023webarena,xie2024osworld,trivedi2024appworld,barres2025tau2}.

\begin{figure}[!t]
    \centering
    \includegraphics[width=\columnwidth]{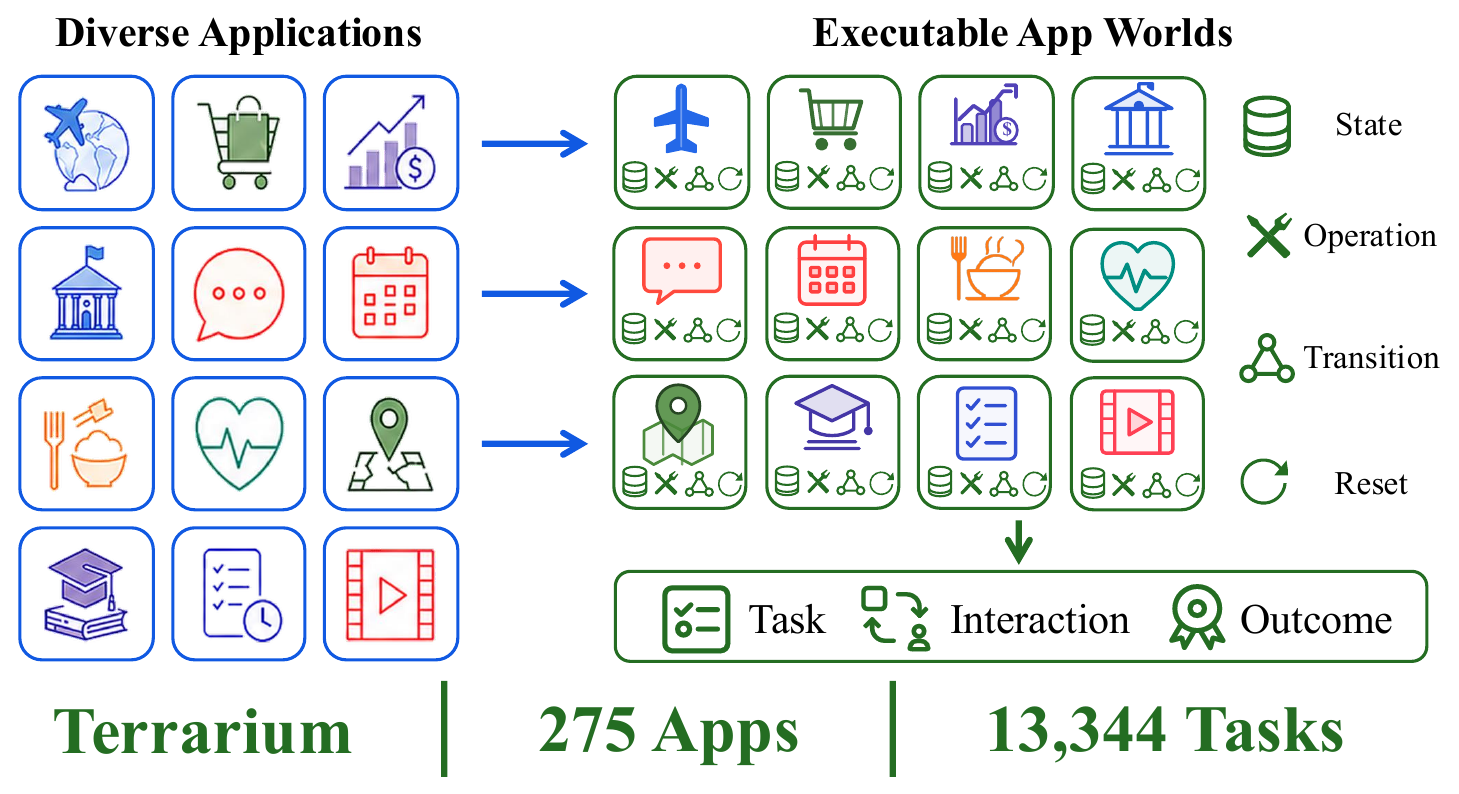}
    \caption{From 275 applications, \toolname{} constructs executable App
    Worlds and 13{,}344 tasks for assistant training.}
    \label{fig:teaser}
\end{figure}

However, constructing environments that are simultaneously realistic,
controllable, and scalable remains challenging.
Real-world applications contain rich workflows, state dynamics, and service
rules, but using them for repeated training is constrained by restricted access
and the difficulty of resetting state
~\citep{garg2025real,chae2026verienv}.
Manually engineered environments~\citep{zhou2023webarena,xie2024osworld,rawles2024androidworld,
trivedi2024appworld} offer greater control, but each additional application requires substantial
implementation and validation effort, making broad ecosystem coverage
difficult.
Automated synthesis can reduce this cost and expand coverage
~\citep{sullivan-etal-2025-procedural,fang2025agentscaler,
song2026envscaler,wang2026awm}.
However, most existing pipelines begin from generic scenarios, often producing
environments that simplify the interactions that LLM agents must ultimately
handle in real-world applications.
This simplification limits the training signal available for developing the
capabilities required for reliable task completion.
These limitations motivate a promising direction that uses rich information
from individual applications to define executable worlds and their training
supervision.

To address this challenge, we introduce \toolname, a scalable framework for app-grounded world and task synthesis.
\toolname{} first synthesizes executable App Worlds from real-world application information.
From these worlds, \toolname{} generates training task packages with evaluator-side rubrics.
To ensure end-to-end consistency across these artifacts, \toolname{} validates them jointly through agent rollouts.
Rollout feedback drives subsequent rounds of refinement and validation, forming a closed-loop construction process.
Figure~\ref{fig:teaser} provides an overview of the executable App Worlds and training tasks constructed by \toolname{} across diverse applications.

\begin{figure*}[!t]
    \centering
    \includegraphics[width=\textwidth]{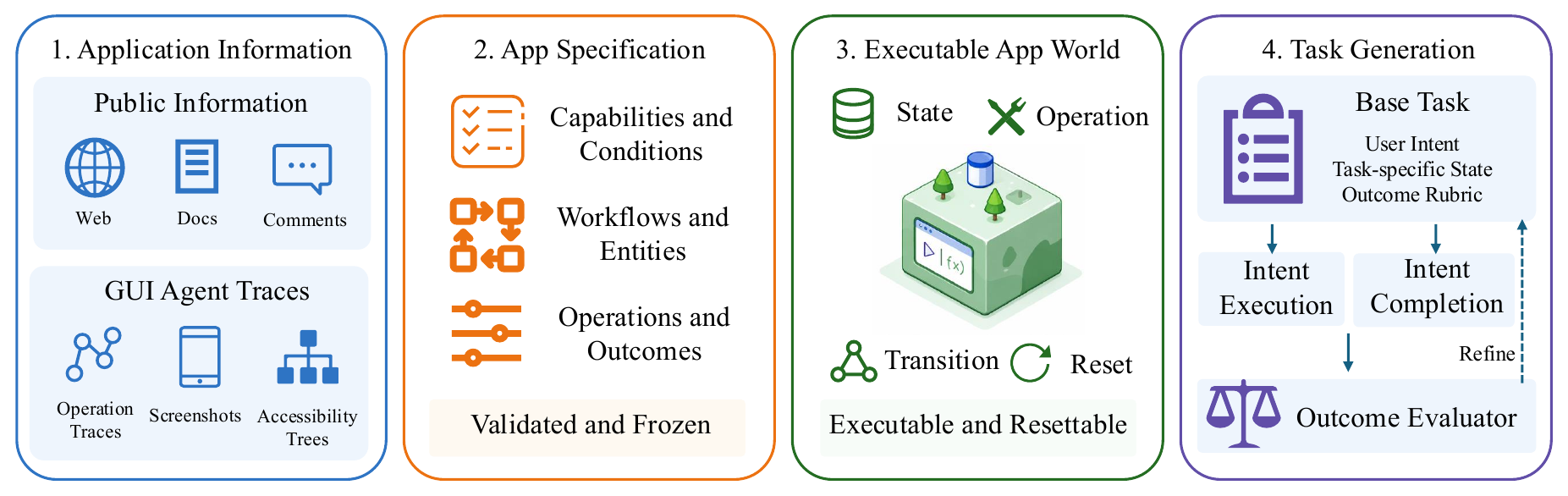}
    \caption{Overview of the \toolname{} construction pipeline.}
    \label{fig:construction-pipeline}
\end{figure*}

The \toolname{} construction pipeline consists of two main stages: App World synthesis and task synthesis (Figure~\ref{fig:construction-pipeline}).
\toolname{} treats real-world application information as a blueprint for App World construction.
It organizes public Web information and GUI-agent traces into structured app specifications that capture application capabilities and conditions, workflows and entities, and operations and observable outcomes.
Once validated and frozen, each specification guides the construction of a self-contained App World with app-specific state, callable operations, transition logic, and a resettable seed state, enabling repeated interaction without access to production backends.
As a common intermediate representation, the app specification supports a shared construction process across diverse applications while retaining app-specific workflows and constraints.
Task synthesis builds on these worlds by grounding user intents in concrete App World states and pairing them with evaluator-side outcome rubrics.
The rubrics define intended outcomes and prohibited side effects without prescribing reference trajectories.
From the resulting base tasks, \toolname{} derives Intent Execution supervision for carrying actionable intents to valid outcomes and Intent Completion supervision for identifying and requesting missing, execution-critical information before changing application state.

We construct a training suite of 275 App Worlds and 13{,}344 training tasks derived from 8{,}878 grounded base tasks.
We train Qwen3 models~\cite{yang2025qwen3technicalreport} at three scales (4B, 8B, and 14B parameters) using Group Relative Policy Optimization~\citep{shao2024deepseekmathpushinglimitsmathematical} with evaluator-side outcome rewards.
Across these scales, models trained with \toolname{} outperform the corresponding Qwen3 models by an average of 18.34 percentage points on the $\tau^2$-bench~\citep{barres2025tau2} and 10.94 percentage points on the AppWorld~\citep{trivedi2024appworld}.
Data-scaling experiments spanning 35 to 275 App Worlds show that both benchmark micro-averages improve consistently as the training corpus grows.
Supervision-composition ablations show that combining Intent Execution and Intent Completion improves over Intent Execution alone across every benchmark subset.
Together, these results show that \toolname{} provides effective supervision across model scales, benefits from broader app-grounded training data, and supports complementary capabilities for executing and completing user intents.

Our main contributions are summarized below.
\begin{enumerate}
    \item \textbf{A scalable framework for App World and task synthesis.}
    We introduce \toolname{}, which transforms real-world application information into executable, resettable App Worlds and grounded training task packages with evaluator-side rubrics.
    Agent rollouts jointly validate and iteratively refine the synthesized worlds, tasks, and evaluators through a closed-loop construction process.

    \item \textbf{An open-source implementation and training suite.}
    We implement \toolname{} at scale and construct a training suite comprising 275 App Worlds and 13{,}344 tasks.
    We will open-source the framework together with the resulting App Worlds and task packages to support reproducible research on training general-purpose application agents.

    \item \textbf{Large-scale training and systematic evaluation.}
    We train Qwen3 models at 4B, 8B, and 14B parameters using GRPO and evaluate them on $\tau^2$-bench and AppWorld.
    Across model scales, \toolname{} yields average gains of 18.34 and 10.94 percentage points, respectively, over the corresponding Qwen3 models.
    Data-scaling experiments and supervision-composition ablations further demonstrate the benefits of expanding the app-grounded training corpus and combining complementary task variants.
\end{enumerate}

\section{Related Work}
\label{sec:related-work}

\subsection{LLM Agents Interacting with Applications}
LLMs are increasingly used as agents that translate user intent into actions within applications. A major line of work enables such interaction
through structured tools and APIs, covering tool invocation, large-scale API retrieval, and planning and execution over real-world
services~\citep{schick2023toolformer,qin2023toolllm,patil2023gorilla,du2024anytool,song2023restgpt}. API-Bank and BFCL complement these methods by
evaluating API selection, argument generation, and multi-step function calling~\citep{li-etal-2023-api,patil2025bfcl}. Another line extends agents
to direct GUI control, including app agents visual understanding and grounding~\citep{hong2024cogagent,cheng-etal-2024-seeclick,zheng2024seeact,deng2023mind2web}, element-
aware GUI testing~\citep{wu2025repeek}, native GUI interaction~\citep{qin2025uitars}, and learning interface dynamics from execution-validated interactions~\citep{wu2026uioceanus,bai2024digirl}. Recent benchmarks further shift from isolated commands to long-horizon tasks involving dynamic
state, domain policies, and users: ToolSandbox evaluates stateful conversational tool use~\citep{lu-etal-2025-toolsandbox}, $\tau$-bench
introduces policy-constrained user conversations~\citep{yao2024tau}, $\tau^2$-bench allows users and agents to modify shared
state~\citep{barres2025tau2}, and MobileWorld combines underspecified instructions, agent--user interaction, and hybrid GUI and MCP
actions~\citep{kong-etal-2026-mobileworld}. These works develop and evaluate
agents in increasingly complex applications. \toolname{} instead focuses on the
training substrate, using Web search results and GUI-agent traces from a particular
real-world application to construct an app specification, an executable App World, and
outcome-based training tasks for Intent Execution and Intent Completion.

\subsection{Training Environments for LLM Agents}

Executable environments provide stateful interaction and outcome-based
feedback for training LLM agents, but constructing them at scale remains
challenging. WebShop, WebArena, VisualWebArena, OSWorld, AndroidWorld, and AppWorld provide executable
web, desktop, mobile, and application environments with controlled state
and execution- or state-based evaluation~\citep{yao2022webshop,zhou2023webarena,
koh2024visualwebarena,xie2024osworld,rawles2024androidworld,trivedi2024appworld}. Because these
environments expose executable state and task-specific evaluators, outcomes can
be checked against the resulting program state. However, application logic,
initial states, and evaluators generally require environment- or task-specific
implementation, making broad application coverage expensive.
Programmatic synthesis seeks to expand this coverage while reducing engineering
effort. These synthesizers differ mainly in their starting points. RandomWorld
samples compositional tools from a manually designed type
system~\citep{sullivan-etal-2025-procedural}. AgentScaler begins from large API
collections~\citep{fang2025agentscaler}. EnvScaler mines environment themes from
existing task sets~\citep{song2026envscaler}, and AWM starts from generic domain
keywords or scenario descriptions~\citep{wang2026awm}; related efforts synthesize trajectories or tasks~\citep{xu2024agenttrek,xie2025agentsynth}. Each then synthesizes
executable state, tools or interfaces, and verification signals with reduced
manual engineering.
The synthesis systems discussed above primarily construct
worlds from type systems, API collections, task-derived themes, or generic
domains. \toolname{} instead combines Web search results with GUI-agent traces
from a particular real-world application, organizes this evidence into
an app specification, and uses the specification to construct both a stateful
App World and its training supervision. This produces app-grounded App Worlds
through a shared pipeline across applications.

\section{\toolname}
\label{sec:method}
\label{sec:world-synthesis}

\subsection{Overview}

\toolname{} converts real-world application information into executable App Worlds and outcome-based training tasks. Its central representation is an app specification that records supported capabilities, workflows, entities, operating conditions, and observable outcomes.
The specification bounds a controllable surrogate of the source application.

Figure~\ref{fig:construction-pipeline} summarizes the construction pipeline.
Web search and GUI-agent traces are consolidated into an app specification, which guides synthesis of an app-specific runtime and a resettable seed world. 
\toolname{} then constructs persona-conditioned base tasks and outcome rubrics, using calibration rollouts to repair inconsistencies among the task, evaluator, and world.
An App World comprises the state schema, model-visible operations, transition logic, and seed state for one application. 
A base task pairs this runtime with a user intent, optional state enrichment, and an evaluator-side rubric. 
\toolname{} constructs Intent Execution and Intent Completion tasks from the base tasks. 
Calibration rollouts evaluate these tasks and trigger revisions to the task, rubric, or App World when construction errors are detected. 
The accepted tasks, evaluators, and App Worlds then provide self-contained artifacts for LLM training.

\begin{table*}[!t]
    \centering
    \small
    \setlength{\tabcolsep}{7pt}
    \newcommand{\resultgain}[1]{\,\textcolor{green!45!black}{(+#1)}}
\caption{Results on $\tau^2$-bench and AppWorld. Parentheses show \toolname{} gains over the corresponding Qwen3 (Thinking) model, and bold marks the best result at each scale.}
    \label{tab:model-scaling}
    \begin{tabular}{@{}lrrrrrrr@{}}
        \toprule
        \multirow{2}{*}{Model} & \multicolumn{4}{c}{$\tau^2$-bench} &
        \multicolumn{3}{c}{AppWorld} \\
        \cmidrule(lr){2-5}\cmidrule(lr){6-8}
        & Airline & Retail & Telecom & Overall &
        Normal & Challenge & Overall \\
        \midrule
        Qwen3-4B (Thinking) & 36.00 & 39.47 & 26.32 & 33.45 & 16.67 & 9.59 & 11.62 \\
        AWM-4B              & 22.00 & 41.23 & 28.07 & 32.37 & 15.48 & 9.35 & 11.11 \\
        EnvScaler-4B        & 44.00 & 58.77 & 28.07 & 43.53 & 15.48 & 12.71 & 13.50 \\
        \textbf{\toolname{}-4B (Ours)} & \textbf{50.00} & \textbf{62.28} & \textbf{54.39} & \textbf{56.83\resultgain{23.38}} & \textbf{30.36} & \textbf{21.10} & \textbf{23.76\resultgain{12.14}} \\
        \midrule
        Qwen3-8B (Thinking) & 34.00 & 46.49 & 37.72 & 40.65 & 23.21 & 16.31 & 18.29 \\
        AWM-8B              & 36.00 & 41.23 & 34.21 & 37.41 & 22.62 & 11.51 & 14.70 \\
        EnvScaler-8B        & 40.00 & 56.14 & 37.72 & 45.68 & 24.40 & 15.83 & 18.29 \\
        \textbf{\toolname{}-8B (Ours)} & \textbf{52.00} & \textbf{64.91} & \textbf{57.89} & \textbf{59.71\resultgain{19.06}} & \textbf{36.90} & \textbf{24.46} & \textbf{28.03\resultgain{9.74}} \\
        \midrule
        Qwen3-14B (Thinking) & 52.00 & 54.39 & 43.86 & 49.64 & 22.62 & 15.83 & 17.78 \\
        AWM-14B               & 34.00 & 39.47 & 38.60 & 38.13 & 29.76 & 16.07 & 20.00 \\
        \textbf{\toolname{}-14B (Ours)} & \textbf{56.00} & \textbf{60.53} & \textbf{66.67} & \textbf{62.23\resultgain{12.59}} & \textbf{36.31} & \textbf{25.66} & \textbf{28.72\resultgain{10.94}} \\
        \bottomrule
    \end{tabular}
\end{table*}

\subsection{From Real-World Application Information to App Specifications}
\label{sec:app-specification}

\paragraph{Application information collection.}
Application-level understanding requires both functional breadth and direct interface grounding. Given an application name, an LLM uses Web search to collect public descriptions, usage instructions, intended users, and operating conditions. A GUI agent then interacts with the application and records action traces, page screenshots, and accessibility trees. Web results identify candidate capabilities, while GUI traces ground them in visible interaction paths, input requirements, and observable outcomes.

\paragraph{App specification construction.}
An LLM consolidates the collected information into a semantic application summary and normalizes each supported capability into workflow steps that describe information gathering, decision points, and application operations.
It then associates these steps with business entities and records how references and values flow through the workflow.
From this entity flow, \toolname{} derives model-visible operation contracts. 
Observation operations expose candidates and constraints, whereas action operations implement atomic state changes. 
Their arguments must be traceable to the user, world state, fixed schema values, or earlier operation returns.
The summary, workflows, entity flow, and operation contracts together form the app specification.

\paragraph{Validation and scope.}
Mechanical checks verify mappings across artifacts, resolvable entity references, and valid argument provenance. 
A capability-level self-check further evaluates whether the documented workflows are executable with information available to the agent. 
When a self-contained world requires details absent from the collected information, \toolname{} may introduce application-typical design choices, which are treated as synthetic world design rather than claims about the production backend. 
The validated specification is then frozen as the semantic contract for world construction.

\subsection{App World Construction}
\label{sec:app-world-construction}

\paragraph{Specification-constrained runtime.}
The app specification defines operational semantics but not an executable world.
\toolname{} synthesizes an app-specific state schema, model-visible operations, and transition logic under this contract. 
App Worlds share a small protocol for initialization, operation invocation, state inspection, and reset, while retaining application-specific schemas and transitions.

\paragraph{Search-guided seed-world construction.}
To populate the runtime, a schema-conditioned search agent collects representative entity instances, value and state distributions, and edge-case patterns from public sources.
Unlike application-level search, this stage is conditioned on a fixed runtime schema and targets data needed to instantiate its business objects.
The agent normalizes the retrieved material into a source pool and derives rules governing entity counts, relations, and state distributions before generating a task-independent seed world.
Individual tasks may add contract-checked state enrichment, while reset restores a fresh seed and prevents cross-episode leakage. 
Calibration may reopen the seed only when a downstream rollout identifies an upstream defect.

\subsection{Task Generation}
\label{sec:task-generation}

Given an executable App World, \toolname{} constructs grounded tasks and their corresponding evaluators, derives two complementary forms of post-training supervision, and validates their joint behavior through rollouts.

\paragraph{Persona-conditioned task authoring.}
To diversify user contexts and expressions, we follow prior work on persona-conditioned task synthesis~\citep{ge2024scaling}.
\toolname{} combines each sampled persona with application capabilities and seed-world information. 
An LLM uses this context to author a plausible use scenario, while the App World continues to determine the executable state and outcome. 
Iterative authoring prioritizes underused scenarios and undercovered application capabilities.

\paragraph{Grounded tasks and evaluators.}
Conditioned on a selected scenario, \toolname{} constructs each base task with a user intent, optional task-specific state enrichment, and an evaluator-side outcome rubric.
The rubric specifies the intended outcome and prohibited side effects without prescribing a reference trajectory. A grounding pass checks that target entities exist, prerequisites can be satisfied, and the outcome is judgeable.
After execution, the evaluator uses the state difference, interaction trajectory, operation returns, and final response to produce a binary reward.
The policy sees only the user intent and model-visible world interface.

\paragraph{Supervision variants.}
From each base task, \toolname{} constructs outcome-based supervision without prescribing a canonical action sequence. 
We call an intent actionable when a rubric-valid execution exists and every execution-critical input is either supplied by the user or recoverable through the model-visible App World.

\paragraph{Intent Execution.}
For an actionable base task, \toolname{} rewrites the user intent as a natural instruction that preserves the requested outcome and application constraints while removing explicit hints about the available operations. 
The App World, episode state, and outcome rubric remain unchanged, and ambiguous rewrites are excluded. 
The evaluator rewards a valid final state and response while allowing any interaction strategy that satisfies the same constraints.

\paragraph{Intent Completion.}
For eligible Intent Execution tasks, \toolname{} removes or obscures one user-supplied field required by the operation contract, transition logic, or outcome rubric. 
The field must be necessary for the original objective and unrecoverable from the remaining intent, valid defaults, or read-only operations. 
The derived rubric rewards a focused clarification question before any state-changing action. 
Guessing the missing value, changing state, or claiming completion results in failure, and the expected state difference remains empty. 
The episode ends with the clarification question without simulating a subsequent user response.

The two variants share the same App World and underlying objective. Intent Execution rewards completing an actionable intent, while Intent Completion rewards eliciting necessary user-only information without changing application state. Requiredness and recoverability are used to construct the tasks and are not exposed to the policy.

\paragraph{Rollout-guided refinement.}
The generated tasks remain provisional until the task, evaluator, and App World operate consistently during execution. 
Static checks cannot verify this joint behavior, so \toolname{} launches a calibration rollout for each provisional task and examines the trajectory, state changes, operation outputs, final response, and evaluator output. 
Failures are classified as policy failures, task or rubric defects, seed or runtime defects, or infrastructure failures. 
Confirmed construction defects revise the responsible artifact, after which the affected task is rerun. 
Policy failures do not modify the task semantics, and infrastructure failures are retried. 
This loop continues until the task, evaluator, and App World pass the closure check or the task is excluded. 
Only tasks that pass this final check are retained as Intent Execution or Intent Completion supervision.

\section{Evaluation}
\label{sec:evaluation}

\subsection{Experimental Setup}
\label{sec:experimental-setup}

\begin{figure}[t]
    \centering
    \includegraphics[width=\columnwidth]{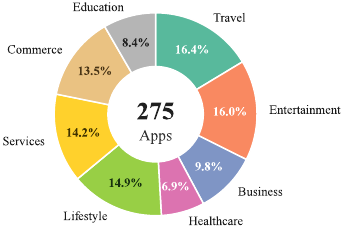}
    \caption{Category distribution of 275 App Worlds.}
    \label{fig:app-category-distribution}
\end{figure}

\paragraph{Training data and models.}
We evaluate whether supervision generated in \toolname{} App Worlds transfers to independently developed application benchmarks.
The final corpus is derived from 8{,}878 grounded base tasks and contains 8{,}867 Intent Execution tasks and 4{,}477 Intent Completion tasks, yielding 13{,}344 tasks in total.
We train 4B, 8B, and 14B Qwen3 Thinking models using GRPO~\citep{shao2024deepseekmathpushinglimitsmathematical} with evaluator-side outcome rewards.
All three model scales use the same configuration: training runs for 340 optimization steps with a learning rate of $7\times10^{-7}$, and each step contains 64 rollout groups with 16 samples per group.
Qwen3.5-397B-A17B~\cite{qwen3.5} serves as the evaluator-side judge, scoring each rollout against its task-specific outcome rubric to produce the reward used by GRPO.
All training runs are implemented with the slime framework~\cite{slime_github} and conducted on 24 NVIDIA H20 GPUs.

\paragraph{Benchmarks and evaluation.}
$\tau^2$-bench~\citep{barres2025tau2} contains 278 tasks across airline,
retail, and telecom services. AppWorld~\citep{trivedi2024appworld} contains 585
tasks across its Normal and Challenge subsets. We report each subset together
with the task-level micro-average over the complete benchmark. We compare each
\toolname{} model with its Qwen3 Thinking counterpart and with
AWM~\citep{wang2026awm} and EnvScaler~\citep{song2026envscaler} at matching
model scales. Invalid model outputs are scored as failures.

\subsection{Main Results}
\label{sec:main-results}

\begin{figure}[t]
    \centering
    \includegraphics[width=\columnwidth]{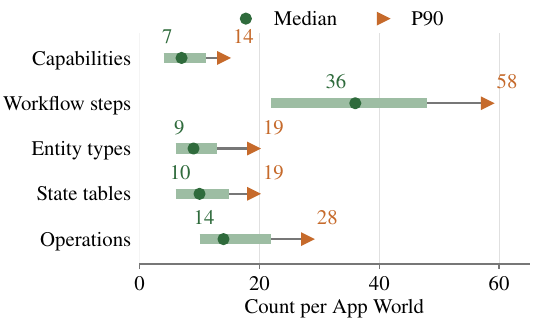}
    \caption{Complexity of 275 App Worlds. Bars show the interquartile range, with markers for the median and P90.}    
    \label{fig:app-data-statistics}
\end{figure}

\toolname{} improves performance consistently across benchmark families and
model scales. As shown in Table~\ref{tab:model-scaling}, it achieves the highest
overall performance on both benchmarks at 4B, 8B, and 14B. Relative to the
corresponding Qwen3 Thinking models, the average improvements are 18.34 points
on $\tau^2$-bench and 10.94 points on AppWorld. Across all three model scales, \toolname{} improves over the corresponding Qwen3 model in every $\tau^2$-bench domain and AppWorld setting. The gains are largest on Telecom, exceeding 20 points at each scale.

The comparison with AWM and EnvScaler shows the same cross-benchmark
consistency. \toolname{}-4B exceeds EnvScaler-4B by 13.30 points on
$\tau^2$-bench and 10.26 points on AppWorld. \toolname{}-8B exceeds
EnvScaler-8B by 14.03 and 9.74 points, while \toolname{}-14B exceeds AWM-14B by
24.10 and 8.72 points. Together, these results demonstrate transfer to both interactive service completion and app-level execution.

Scaling \toolname{} from 4B to 14B improves the micro-average by 5.40 points on $\tau^2$-bench and 4.96 points on AppWorld. The $\tau^2$-bench
micro-average rises from 56.83 at 4B to 59.71 at 8B and 62.23 at 14B. AppWorld
improves from 23.76 to 28.03 and 28.72, with most of the increase occurring
between 4B and 8B. At the subset level, Telecom and AppWorld Challenge continue
to improve through 14B, whereas Retail peaks at 8B and AppWorld Normal changes
only slightly from 8B to 14B.

\begin{table*}[!t]
    \centering
    \small
    \caption{Ablations of training strategy and supervision design on Qwen3-4B. Panels A and B compare training strategies and supervision variants, respectively.}
\label{tab:training-supervision-ablation}
    \begin{tabular*}{\textwidth}{@{\extracolsep{\fill}}lrrrrrrr@{}}
        \toprule
        \multirow{2}{*}{Configuration} & \multicolumn{4}{c}{$\tau^2$-bench} & \multicolumn{3}{c}{AppWorld} \\
        \cmidrule(lr){2-5}\cmidrule(lr){6-8}
         & Airline & Retail & Telecom & Overall & Normal & Challenge & Overall \\
        \midrule
        \multicolumn{8}{@{}l}{\textbf{A. Training strategy}} \\
        Qwen3-4B (Thinking)       & 36.00 & 39.47 & 26.32 & 33.45 & 16.67 & 9.59 & 11.62 \\
        SFT on accepted rollouts & 28.00 & 40.35 & 5.26 & 23.74 & 26.79 & 13.67 & 17.44 \\
        SFT on all rollouts        & 36.00 & 49.12 & 8.77 & 30.22 & \textbf{31.55} & 15.83 & 20.34 \\
        GRPO          & \textbf{50.00} & \textbf{62.28} & \textbf{54.39} & \textbf{56.83} & 30.36 & \textbf{21.10} & \textbf{23.76} \\
        \midrule
        \multicolumn{8}{@{}l}{\textbf{B. Supervision design}} \\
        Qwen3-4B (Thinking)                         & 36.00 & 39.47 & 26.32 & 33.45 & 16.67 & 9.59 & 11.62 \\
        Execution only                              & 34.00 & 44.74 & 36.84 & 39.57 & 26.79 & 18.23 & 20.68 \\
        Execution + Completion + Handoff            & \textbf{54.00} & 57.89 & \textbf{57.02} & \textbf{56.83} & 27.98 & 17.75 & 20.68 \\
        Execution + Completion + Confirmation       & 44.00 & 57.89 & 52.63 & 53.24 & 26.19 & 20.38 & 22.05 \\
        Execution + Completion                      & 50.00 & \textbf{62.28} & 54.39 & \textbf{56.83} & \textbf{30.36} & \textbf{21.10} & \textbf{23.76} \\
        \bottomrule
    \end{tabular*}
\end{table*}

\subsection{Diversity and Complexity of Synthesized App Worlds}
\label{sec:app-world-structure}

\begin{figure}[t]
    \centering
    \includegraphics[width=\columnwidth]{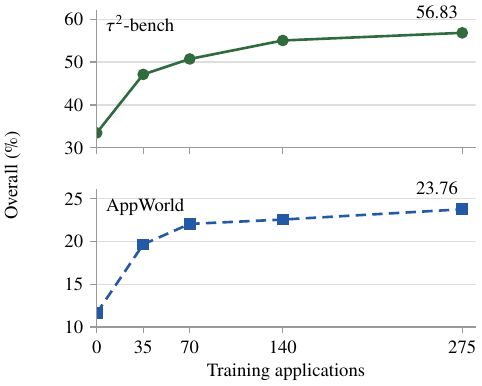}
    \caption{Performance versus the number of training App Worlds on $\tau^2$-bench and AppWorld.}
    \label{fig:data-scaling}
\end{figure}

Figure~\ref{fig:app-category-distribution} shows that the App Worlds cover a broad range of application categories. The 275 worlds span 8 categories, each accounting for 6.9\%--16.4\% of the corpus. This coverage broadens the application contexts represented during task generation, including differences in user goals, workflows, and operating conditions.

We characterize the structural complexity of the synthesized App Worlds using functional breadth, workflow depth, state structure, and available operations. Capabilities measure functional breadth, and workflow
steps measure procedural depth. Entity types and state tables characterize the
represented business state, while agent-visible operations define the interface
through which a model can observe and change that state.
Figure~\ref{fig:app-data-statistics} reports the corresponding per-world
distributions over the final set of 275 accepted App Worlds.

Figure~\ref{fig:app-data-statistics} reveals two properties of the synthesized
corpus. First, substantial structure is present in a typical App World. The
median world combines 7 capabilities and 36 documented workflow steps with 9
entity types, 10 state tables, and 14 agent-visible operations. This combination
provides the procedural, state, and interface structure needed for multi-step
interaction and state-based outcome evaluation. Second, the
corpus spans a broad range of structural scale. At P90, capabilities and
operations both double relative to the median, entity types and state tables
increase to 19, and workflow steps increase to 58. The upper tail across all 4
dimensions indicates that the synthesis process produces worlds of varying
scope and avoids collapse to a uniform shallow template. These
measurements establish the structural breadth of the corpus, while the external
evaluations examine whether supervision from these worlds transfers beyond the
synthesized environments.

\subsection{Scaling App-Grounded Training Data}
\label{sec:data-scaling}

We study whether transfer improves as the app-grounded training corpus scales. We construct nested corpora containing 35, 70, 140, and 275 App Worlds by randomly sampling App Worlds without replacement, with each larger corpus retaining all worlds from the preceding smaller corpus.
We train Qwen3-4B on each corpus using the same GRPO configuration. The Qwen3-4B
(Thinking) model serves as the zero-world reference. 

Figure~\ref{fig:data-scaling} shows that both benchmark micro-averages improve consistently as the app-grounded training corpus expands.
Relative to the zero-world reference, training on 35 App Worlds improves performance by 13.67 percentage points on $\tau^2$-bench and 8.04 percentage points on AppWorld.
Across the trained models, increasing the corpus from 35 to 275 App Worlds raises performance from 47.12 to 56.83 on $\tau^2$-bench and from 19.66 to 23.76 on AppWorld.
These continued gains suggest that scaling app-grounded environments and their associated supervision is a promising direction for further improving cross-application transfer.

\subsection{Training Strategy and Supervision Design}
\label{sec:training-supervision-ablation}

Table~\ref{tab:training-supervision-ablation} evaluates training strategies and supervision compositions on Qwen3-4B.
Panel A compares trajectory-based SFT with outcome-based GRPO.
For SFT data collection, we use Qwen3.5-397B-A17B to generate trajectories on the synthesized training tasks.
SFT on accepted rollouts retains only trajectories that satisfy the task evaluator, 
whereas SFT on all rollouts retains the complete collection regardless of evaluation outcome.
Outcome-based GRPO instead generates on-policy rollouts from the model being trained and uses the evaluator result as a binary reward.
Panel B applies GRPO and compares supervision mixtures constructed from Intent Execution, Intent Completion, Handoff, and Confirmation tasks. Handoff and Confirmation are auxiliary task variants derived from grounded base tasks by modifying the task conditions and evaluator. A Handoff task requires the agent to complete the operations supported by the App World and return the remaining user-side, device-side, offline, or external-system step to the user. A Confirmation task provides enough information to determine a state-changing operation but requires the agent to request user approval without executing that operation.

\paragraph{Training strategy.}
GRPO is the only evaluated training strategy that improves over the unmodified Qwen3-4B model on both benchmarks, with gains of 23.38 percentage points on $\tau^2$-bench and 12.14 percentage points on AppWorld.
For SFT, retaining all collected rollouts rather than only accepted ones improves performance by 6.48 points on $\tau^2$-bench and 2.90 points on AppWorld.
Nevertheless, both SFT variants remain below the base model on $\tau^2$-bench despite improving on AppWorld.    

The difference between GRPO and SFT is clearest on more difficult and out-of-distribution tasks. SFT on all rollouts performs well on Retail and AppWorld Normal but fails to maintain this performance on Telecom and AppWorld Challenge, falling to 8.77 and 15.83, respectively.
Telecom introduces a dual-control setting in which both the user and the agent can modify environment state~\citep{barres2025tau2}, while AppWorld Challenge contains harder tasks.
These settings introduce stronger shifts in interaction structure and task difficulty.
GRPO reaches 54.39 on Telecom and 21.10 on AppWorld Challenge, suggesting that GRPO provides more robust transfer under distribution shift than trajectory-based SFT.

\paragraph{Intent Completion supervision.}
Combining Intent Completion with Intent Execution improves every benchmark subset.
The combination raises the $\tau^2$-bench from 39.57 to 56.83 and the AppWorld from 20.68 to 23.76.
The larger improvement on $\tau^2$-bench is consistent with its interactive task structure, where successful execution often depends on identifying and eliciting information available only from the user.
The gain on AppWorld, which primarily evaluates completed application state, suggests that the benefit extends beyond clarification itself: learning to recognize missing prerequisites and avoid premature state changes also supports more reliable execution.
These results demonstrate that Intent Execution and Intent Completion provide complementary supervision for acting on complete requests and handling incomplete ones.

\paragraph{Auxiliary task variants.}
Handoff and Confirmation introduce different behavioral trade-offs rather than uniform improvements.
Including Handoff reduces AppWorld by 3.08 points, improves Airline and Telecom by 4.00 and 2.63 points while reducing Retail by 4.39 points.
This pattern suggests that Handoff can help when tasks cross agent, user, or system boundaries, but may also encourage transferring responsibility when the benchmark expects the agent to complete the task within the application.
Including Confirmation reduces the $\tau^2$-bench and AppWorld by 3.59 and 1.71 points, respectively.
Because Confirmation rewards requesting approval before a state-changing operation, it can conflict with evaluation tasks that expect immediate execution of an already specified request.
Overall, neither auxiliary variant improves both benchmarks, indicating that Handoff and Confirmation encode narrower behaviors whose value depends on whether a task actually requires delegation or approval.
Among the evaluated mixtures, Intent Execution and Intent Completion provide the strongest balance between completing actionable requests and eliciting information when execution is not yet possible.  
\section{Conclusion}
\label{sec:conclusion}

We introduced \toolname, a scalable framework for app-grounded world and task synthesis to support training general-purpose assistants across broad application ecosystems.
\toolname{} transforms real-world application information into executable, resettable App Worlds and grounded task packages with evaluator-side rubrics, using agent rollouts for closed-loop validation and refinement.
We implement \toolname{} at scale to construct 275 App Worlds and 13{,}344 training tasks, and will open-source the framework and resulting artifacts.
GRPO training across Qwen3 models with 4B, 8B, and 14B parameters yields average gains of 18.34 percentage points on $\tau^2$-bench and 10.94 percentage points on AppWorld over the corresponding Qwen3 models.
Data-scaling and supervision-composition studies further demonstrate the benefits of broader app-grounded training data and complementary task variants.
Together, these results suggest that app-grounded world and task synthesis offers a scalable path toward general-purpose assistants capable of operating across diverse applications.

\clearpage
\bibliography{references}

\clearpage
\appendix
\section{Full Training Configuration}
\label{supp:training}

\subsection{Models, objective, and supervision sampling}

The final models are trained with Group Relative Policy Optimization
(GRPO)~\citep{shao2024deepseekmathpushinglimitsmathematical}, using a binary
task-outcome reward. We initialize 4B, 8B, and 14B Qwen3 Thinking models
~\citep{yang2025qwen3technicalreport}. The same optimization recipe is used at
all three model scales:

\begin{itemize}
  \item \textbf{Model scales:} Qwen3 Thinking 4B, 8B, and 14B.
  \item \textbf{Objective:} GRPO with a binary task-outcome reward.
  \item \textbf{Learning rate and duration:} $7\times10^{-7}$ for 340 update
  steps.
  \item \textbf{Rollout batch:} 64 groups per step and 16 samples per group,
  yielding 1,024 rollout samples per step.
  \item \textbf{Supervision sampling:} Intent Execution and Intent Completion
  are sampled at a 70/30 ratio.
  \item \textbf{Infrastructure:} the slime framework~\citep{slime_github} on
  24 NVIDIA H20 GPUs.
\end{itemize}

The primary corpus contains 8,867 Intent Execution tasks and 4,477 Intent
Completion tasks. Thus, the stored task-count proportions are 66.45\% and
33.55\%, respectively.

\subsection{Outcome judge and reward construction}

Qwen3.5-397B-A17B-FP8~\citep{qwen3.5} serves as the evaluator-side judge during
GRPO. For each rollout, the evaluator is given the task-specific outcome rubric,
the pre/post business-state difference, the interaction trajectory, operation
returns, and the assistant's final response. The rubric defines acceptable
outcomes and prohibited side effects without prescribing a canonical operation
sequence. The parsed verdict is converted to a binary reward: a valid pass
receives one and a valid failure receives zero.

The policy sees only the user intent and model-visible operation schemas and
returns. It does not see the evaluator rubric, structured expected evidence, or
reference traces. Expected-evidence records may specify acceptable final facts,
state-difference shapes, and forbidden states, but may not contain a required
tool sequence or step-by-step solution.

\paragraph{Judge prompt and input.}
The judge receives the following system prompt (line breaks are added only for
typesetting):
\begingroup\small
\begin{verbatim}
You are an evaluator for a miniapp
agent episode. Judge only the final
outcome using the rubric and evidence.
Do not require a fixed tool sequence.
DB diff and actual tool returns are
primary evidence for committed business
writes. Do not treat offers, future
suggestions, or clarification questions
as committed state changes unless a tool
call, DB diff, or final answer explicitly
says the action has already been
completed. Return strict JSON only.
\end{verbatim}
\endgroup
The corresponding user-prompt template is:
\begingroup\small
\begin{verbatim}
Evaluate this episode. Return JSON with
keys: passed (boolean), reward (number),
rationale (string), failed_criteria
(array of strings), failure_classification
(string or null). Reward must be 1.0 for
pass and 0.0 for fail. Allowed
failure_classification values for failed
episodes: agent_failure, task_query_issue,
rubric_or_judge_issue,
expected_evidence_issue,
seed_enrichment_issue,
backend_behavior_issue,
tool_schema_or_description_issue,
runtime_or_infra_issue.

{JUDGE_INPUT_JSON}
\end{verbatim}
\endgroup
The placeholder is replaced by an indented JSON serialization with the
following schema:
\begingroup\small
\begin{verbatim}
{
  "task_id": "...",
  "env_id": "...",
  "user_query": "...",
  "rubric": "...",
  "metadata": {},
  "expected_evidence": {},
  "compact_db_diff": {},
  "full_db_diff": {},
  "db_diff_metadata": {},
  "trajectory": [],
  "tool_returns": [],
  "final_answer": "..."
}
\end{verbatim}
\endgroup
Judge decoding uses temperature 0, a 4,096-token generation limit, and thinking
disabled. Although the response schema includes a numeric reward, the training
pipeline recomputes it from the parsed \texttt{passed} field, assigning 1.0 to a
pass and 0.0 to a failure.

\section{Benchmark, Baseline, and Transfer Protocol}
\label{supp:evaluation-protocol}

\subsection{Benchmarks and score computation}

We evaluate transfer on two independently developed benchmarks.
$\tau^2$-bench contains 278 tasks in Airline, Retail, and Telecom
~\citep{barres2025tau2}. AppWorld contains 585 tasks in its Normal and Challenge
subsets~\citep{trivedi2024appworld}. Subset scores are percentages of tasks
passed. ``Overall'' is the task-level micro-average. Invalid model outputs are
scored as failures.

Official benchmark evaluators produce all reported scores. The Qwen3.5 judge in
Section~\ref{supp:training} provides GRPO rewards and calibration judgments.

Evaluation follows each benchmark's official harness. $\tau^2$-bench uses a
model-based user simulator, while AppWorld does not use a user simulator.
Both the $\tau^2$-bench user simulator and judge use
Qwen3.5-397B-A17B; all other settings follow the default benchmark
configuration.

All reported results use one trained checkpoint and one evaluation run per
configuration.

\subsection{Baselines}

We compare with the corresponding Qwen3 Thinking checkpoint and with AWM
~\citep{wang2026awm} and EnvScaler~\citep{song2026envscaler} at matching model
scales where available.

\subsection{Transfer boundary and overlap audit}

The 275 training applications were sampled from an existing mini-program
collection without reference to either evaluation benchmark. Benchmark task
files, reference answers, and evaluator labels were not used as direct inputs
to App World or task synthesis. The training and evaluation suites may cover
common application domains, but their environments and task instances were
developed independently.

\begin{table*}[t]
  \centering
  \small
  \setlength{\tabcolsep}{5pt}
  \caption{Calibration failure classification and routing.}
  \label{tab:failure-routing}
  \begin{tabular}{@{}p{0.19\textwidth}p{0.32\textwidth}p{0.39\textwidth}@{}}
    \toprule
    Failure class & Classification rule & Required action \\
    \midrule
    Policy failure
      & The task, tool schema, seed, runtime, rubric, evidence, and judge wiring are coherent, but the rollout does not satisfy the outcome
      & Retain as a policy-failure example \\
    Task/rubric defect
      & The query is ambiguous or infeasible, the rubric is too strict/loose, the evidence targets the wrong object or state, or process requirements replace outcome requirements
      & Repair the query, rubric, expected evidence, or judge-input construction; rerun every affected task \\
    Seed/runtime defect
      & Required entities or relations are absent, enrichment is inconsistent, a valid transition is impossible, the tool contract disagrees with implementation, or reset/diff is incorrect
      & Repair the seed, enrichment, operation schema, transition, or state-diff provider; reset and rerun affected tasks \\
    Infrastructure failure
      & API, runner, adapter, serialization, timeout, path, or judge-schema failure occurs independently of task semantics
      & Repair or retry infrastructure; exclude the episode from task difficulty and outcome statistics \\
    \bottomrule
  \end{tabular}
\end{table*}

\section{Closed-Loop Validation and Refinement}
\label{supp:closure}

\subsection{Why rollout closure is needed}

Static checks can validate schemas and references but cannot establish that a
user intent, task-specific seed, runtime, and evaluator behave consistently
during execution. Each provisional task is therefore executed in a calibration
episode. The review jointly inspects the policy trajectory, operation returns,
final response, business-state difference, and evaluator verdict. A defect is
routed back to the artifact responsible for the inconsistency, and the affected
task is rerun after repair.

\subsection{Failure classification and repair routing}

Table~\ref{tab:failure-routing} defines the failure taxonomy and repair route.

\subsection{Closure condition}

A task reaches closure only after all of the following hold:
\begin{enumerate}
  \setlength{\itemsep}{0pt}
  \setlength{\parskip}{0pt}
  \item its package is mechanically valid and the runtime can import, reset,
        expose the model-visible tools, execute operations, finish, and emit a
        serializable state difference;
  \item the query, seed enrichment, operation contracts, rubric, and expected
        evidence refer to consistent entities, states, and boundaries;
  \item any task-, evaluator-, seed-, runtime-, or infrastructure-side issue
        found in rollout review has been repaired and the affected task rerun;
  \item the final reviewed outcome is either a pass or a policy-side failure;
        and
  \item no unresolved construction- or infrastructure-side failure remains.
\end{enumerate}

Infrastructure retries do not count as semantic refinement. A confirmed
construction repair increments the task's refinement history. Tasks that exceed
the allowed repair policy are excluded.

\section{Dataset Card and Complete Statistics}
\label{supp:dataset-card}

\subsection{Dataset identity, intended use, and provenance}

The final snapshot contains 275 accepted App Worlds with anonymized identifiers
\texttt{app\_001} through \texttt{app\_275}.

The dataset is intended for research on language-model interaction with
resettable application surrogates and evaluator-side outcome supervision. An App
World is a synthetic, controllable surrogate constructed from public application
information.

\subsection{Task and supervision-view counts}

A grounded base task defines one executable objective, optional state
enrichment, and evaluator-side outcome boundary. Supervision views reuse that
world and underlying objective.

The primary training pool contains the 8,867 Execution and 4,477 Completion
views, or 13,344 stored tasks. Confirmation and Handoff are auxiliary views used
for ablations and are not included in that total.

\subsection{Complete structural statistics}

Table~\ref{tab:complete-stats} reports mean, P25, median, P75, P90, and maximum.
Except for workflow steps, the statistical unit is one App World. Workflow-step
arrays are flattened over all 2,172 capabilities before aggregation. All
percentiles use the nearest-rank rule.

\begin{table*}[t]
  \centering
  \small
  \setlength{\tabcolsep}{6pt}
  \caption{Complete structural profile of the final accepted corpus.}
  \label{tab:complete-stats}
  \begin{tabular}{@{}lrrrrrr@{}}
    \toprule
    Quantity & Mean & P25 & Median & P75 & P90 & Maximum \\
    \midrule
    Capabilities / world & 7.90 & 4 & 7 & 11 & 14 & 26 \\
    Steps / capability & 4.60 & 4 & 4 & 5 & 7 & 13 \\
    Entity types / world & 10.21 & 6 & 9 & 13 & 19 & 40 \\
    State tables / world & 11.11 & 6 & 10 & 15 & 19 & 40 \\
    Operations / world & 15.81 & 10 & 14 & 22 & 28 & 46 \\
    \quad Observation & 9.62 & 5 & 8 & 13 & 17 & 29 \\
    \quad Action & 6.20 & 3 & 5 & 9 & 12 & 21 \\
    Base tasks / world & 32.28 & 20 & 28 & 40 & 52 & 140 \\
    Execution views / world & 32.24 & 20 & 28 & 40 & 52 & 140 \\
    Completion views / world & 16.28 & 10 & 15 & 20 & 26 & 76 \\
    Confirmation views / world & 11.99 & 6 & 10 & 16 & 22 & 52 \\
    Handoff views / world & 9.97 & 5 & 9 & 14 & 20 & 38 \\
    Execution + Completion / world & 48.52 & 34 & 44 & 57 & 77 & 168 \\
    \bottomrule
  \end{tabular}
\end{table*}

The corresponding integrity totals are 2,172 capabilities, 9,999
capability-level workflow steps, 2,808 entity types, 3,054 business-state
tables, 2,645 observation operations, and 1,704 action operations.
Business-state tables are defined by each runtime's snapshot/diff contract. The
count uses explicit business-table declarations for 230 worlds, explicit
snapshot collections for 38, and schema enumeration excluding app-specific
infrastructure tables for 7.

\subsection{Domain categories}

Table~\ref{tab:category-map} presents the eight domain categories used in the
main paper and a brief description of each.

\begin{table*}[t]
  \centering
  \small
  \setlength{\tabcolsep}{6pt}
  \caption{The eight App World categories used in the main paper and their
  descriptions. Counts and percentages are over 275 App Worlds.}
  \label{tab:category-map}
  \begin{tabular}{@{}lrp{0.62\textwidth}@{}}
    \toprule
    Category & Count (\%) & Category description \\
    \midrule
    Travel & 45 (16.4) & Mobility, trip planning, transportation, and logistics. \\
    Entertainment & 44 (16.0) & Media, sports, events, and other leisure activities. \\
    Commerce & 37 (13.5) & Shopping, retail, and related commercial transactions. \\
    Services & 39 (14.2) & Government and public services, utilities, and information access. \\
    Lifestyle & 41 (14.9) & Food, hospitality, local services, and social or community activities. \\
    Business & 27 (9.8) & Workplace productivity, business operations, finance, and payments. \\
    Healthcare & 19 (6.9) & Health services, care coordination, and related administration. \\
    Education & 23 (8.4) & Learning, educational services, and academic administration. \\
    \bottomrule
  \end{tabular}
\end{table*}

\section{Scaling and Ablation Reproducibility}
\label{supp:scaling-ablation}

\subsection{Data-scaling construction}

The scaling study uses one nested sequence of 35, 70, 140, and 275 App Worlds,
sampled without replacement so that every larger corpus contains the preceding
smaller corpus. Qwen3-4B is trained with the same reported GRPO recipe at every
nonzero scale; the zero-world row is the unmodified Qwen3-4B Thinking
checkpoint. Every nonzero scale uses 340 steps, 64 groups per step, and 16
samples per group.

The 35-world corpus contains 1,237 Intent Execution tasks (73.6\%) and
444 Intent Completion tasks (26.4\%), for 1,681 tasks in total. The 70-world
corpus contains 2,396 Execution tasks (71.6\%) and 950 Completion tasks
(28.4\%), for 3,346 tasks in total. The 140-world corpus contains 4,720
Execution tasks (71.0\%) and 1,928 Completion tasks (29.0\%), for 6,648 tasks
in total. The full 275-world corpus contains 8,867 Execution tasks (66.45\%)
and 4,477 Completion tasks (33.55\%), for 13,344 tasks in total.

\subsection{Training-strategy ablation}

Both SFT configurations are constructed from the same pool of 12,571
trajectories generated by Qwen3.5-397B-A17B-FP8 on the synthesized tasks. The
teacher runs with thinking enabled, preserves its reasoning and interaction
history, and uses temperature 1.0, top-$p$ 1.0, no top-$k$ restriction, a
4,096-token per-turn generation limit, and at most eight turns. The same model
also provides the task-outcome judgment. ``SFT on accepted rollouts'' retains
7,362 judge-accepted trajectories, whereas ``SFT on all rollouts'' retains
12,498 trajectories, comprising the same 7,362 accepted examples and 5,136
non-accepted examples. Conversations longer than 32,768 tokens are discarded
as complete units rather than truncated, preserving assistant--tool interaction
structure.

Both variants perform full-parameter BF16 training of Qwen3-4B for three epochs
with a global batch size of 128 sequences and a maximum sequence length of
32,768. We apply assistant-only per-token loss to every assistant turn in the
preserved multi-turn history. Assistant reasoning, structured tool calls, and
the turn-ending token are supervised, while system, user, and tool-return tokens
are masked. Training uses distributed Adam with peak learning rate
$1\times10^{-5}$, minimum learning rate $1\times10^{-6}$, cosine decay, 10\%
warmup, $\beta_1=0.9$, $\beta_2=0.95$, $\epsilon=10^{-8}$, weight decay 0.1,
and gradient clipping at 1.0. Each run uses the slime
framework~\citep{slime_github} on eight GPUs with tensor, pipeline, context,
and data parallel sizes of 1, 1, 4, and 2, respectively. Data shuffling uses
seed 42 and model training uses seed 1234.

Incomplete tail batches are dropped at each epoch. Accepted-only SFT therefore
performs 57 updates per epoch and 171 updates in total, while all-outcomes SFT
performs 97 updates per epoch and 291 updates in total. We evaluate the final
epoch-three checkpoint; no validation-based checkpoint selection is performed.
Although the two configurations share the model, optimizer, batch size, epoch
count, and training implementation, retaining all outcomes also increases the
number of training examples, supervised tokens, and optimization updates.
GRPO instead collects on-policy rollouts and optimizes the evaluator reward.

\subsection{BFCL evaluation}
\label{supp:bfcl}

We additionally evaluate the base models, \toolname{} models, ablation
checkpoints, and external baselines on the BFCL Multi-Turn suite
~\citep{patil2025bfcl}. Each of its Base, Missing Function, Missing Parameter,
and Long Context splits contains 200 examples. Overall is the macro-average
across the four splits. All models are evaluated with thinking enabled and
thinking history preserved.

\begin{table*}[t]
  \centering
  \small
  \setlength{\tabcolsep}{5pt}
  \caption{BFCL Multi-Turn accuracy (\%). Each split contains 200 examples,
  and Overall is the macro-average across Base, Missing Function, Missing
  Parameter, and Long Context. All models are evaluated with thinking enabled
  and thinking history preserved.}
  \label{tab:bfcl-results}
  \begin{tabular}{@{}lrrrrr@{}}
    \toprule
    Model & Base & Miss. Func. & Miss. Param. & Long Context & Overall \\
    \midrule
    \multicolumn{6}{@{}l}{\textbf{Unmodified Qwen3}} \\
    Qwen3-4B  & 39.00 & 28.00 & 32.00 & 21.00 & 30.00 \\
    Qwen3-8B  & 45.00 & 43.00 & 36.00 & 23.50 & 36.88 \\
    Qwen3-14B & 50.50 & 37.00 & 36.50 & 29.50 & 38.38 \\
    \midrule
    \multicolumn{6}{@{}l}{\textbf{Data scaling}} \\
    \toolname{}-4B (35 App Worlds)  & 47.00 & 27.50 & 36.50 & 29.00 & 35.00 \\
    \toolname{}-4B (70 App Worlds)  & 53.00 & 38.00 & 42.50 & 28.50 & 40.50 \\
    \toolname{}-4B (140 App Worlds) & 48.50 & 37.50 & 46.00 & 29.50 & 40.38 \\
    \toolname{}-4B (275 App Worlds) & 47.50 & 33.50 & 41.50 & 30.50 & 38.25 \\
    \midrule
    \multicolumn{6}{@{}l}{\textbf{Model scaling}} \\
    \toolname{}-8B (275 App Worlds)  & 54.00 & 49.00 & 46.00 & 31.00 & 45.00 \\
    \toolname{}-14B (275 App Worlds) & 54.50 & 44.50 & 43.50 & 31.50 & 43.50 \\
    \midrule
    \multicolumn{6}{@{}l}{\textbf{Supervision-data ablation}} \\
    \toolname{}-4B (Execution only) & 60.00 & 41.00 & 38.00 & 40.50 & 44.88 \\
    \toolname{}-4B (Execution + Completion + Handoff) & 46.00 & 30.00 & 38.00 & 28.50 & 35.63 \\
    \toolname{}-4B (Execution + Completion + Confirmation) & 46.50 & 30.00 & 38.00 & 26.00 & 35.13 \\
    \midrule
    \multicolumn{6}{@{}l}{\textbf{Training-strategy ablation}} \\
    SFT on accepted rollouts (epoch 3) & 13.50 & 14.50 & 9.50 & 6.50 & 11.00 \\
    SFT on all rollouts (epoch 3) & 18.50 & 16.50 & 11.00 & 12.00 & 14.50 \\
    \midrule
    \multicolumn{6}{@{}l}{\textbf{External baselines}} \\
    AWM-4B & 30.00 & 29.00 & 28.50 & 16.50 & 26.00 \\
    AWM-8B & 40.50 & 38.50 & 35.50 & 25.50 & 35.00 \\
    AWM-14B & 48.50 & 36.50 & 36.50 & 28.50 & 37.50 \\
    EnvScaler-4B & 58.50 & 39.50 & 36.50 & 32.50 & 41.75 \\
    EnvScaler-8B & 65.00 & 45.00 & 40.00 & 38.00 & 47.00 \\
    \bottomrule
  \end{tabular}
\end{table*}

\end{document}